\documentclass[11pt]{article}

\usepackage[preprint]{acl}

\usepackage{times}
\usepackage{latexsym}
\usepackage[T1]{fontenc}
\usepackage[utf8]{inputenc}
\usepackage{microtype}
\usepackage{graphicx}
\usepackage{booktabs}
\usepackage{amsmath}
\usepackage{multirow}
\usepackage{xcolor}
\usepackage{adjustbox}

\title{A Systematic Comparison of Prompting and Multi-Agent Methods \\ for LLM-based Stance Detection}

\author{
\textbf{Genan Dai}, 
\textbf{Zini Chen}, 
\textbf{Yi Yang}, 
\textbf{Bowen Zhang}\\  
  School of Artificial Intelligence, Shenzhen Technology University, Shenzhen, China\\
  }
\begin{document}
\maketitle

\begin{abstract}
Stance detection identifies the attitude of a text author toward a given target.
Recent studies have explored various LLM-based strategies for this task, from zero-shot prompting to multi-agent debate.
However, existing works differ in data splits, base models, and evaluation protocols, making fair comparison difficult.
We conduct a systematic comparison that evaluates five methods across two categories---prompt-based inference (Direct Prompting, Auto-CoT, StSQA) and agent-based debate (COLA, MPRF)---on four datasets with 14 subtasks, using 15 LLMs from six model families with parameter sizes from 7B to 72B+.
Our experiments yield several findings.
First, on all models with complete results, the best prompt-based method outperforms the best agent-based method, while agent methods require 7 to 12 times more API calls per sample.
Second, model scale has a larger impact on performance than method choice, with gains plateauing around 32B.
Third, reasoning-enhanced models (DeepSeek-R1) do not consistently outperform general models of the same size on this task.
We release all code, prompts, and per-target results to facilitate future research.
% \footnote{\url{https://github.com/xxx/stance-eval}}
\end{abstract}

%=============================================
\section{Introduction}
%=============================================

Stance detection identifies whether a text author is in favor of, against, or neutral toward a specific target \cite{kuccuk2020stance, zhang2024survey}.
Unlike general sentiment analysis, this task focuses on the attitude toward a particular target, even when the text does not explicitly mention it \cite{mohammad2016semeval}.

With the development of large language models (LLMs), researchers have explored various approaches for stance detection.
These approaches fall into two categories.
The first is prompt-based methods, where a single LLM call with a carefully designed prompt predicts the stance \cite{zhang2022chatgpt, zhang2023stsqa, zhang2022autocot}.
The second is agent-based methods, where multiple LLM agents collaborate through debate and reasoning \cite{lan2024cola, zhangmprf2025}.

However, a critical problem is the lack of unified experimental settings.
As Table~\ref{tab:inconsistency} shows, existing works use different data splits, base LLMs, and evaluation details.
Some evaluate on the full dataset while others use only the test split.
Some use GPT-3.5 while others use GPT-4 or open-source models.
This makes it impossible to fairly compare different methods.

\begin{table}[t]
\centering
\small
\begin{tabular}{lccc}
\toprule
\textbf{Work} & \textbf{Split} & \textbf{Base LLM} & \textbf{Metric} \\
\midrule
\citet{zhang2022chatgpt} & Full & ChatGPT & $F_{avg}$ \\
\citet{lan2024cola} & Test & GPT-3.5 & $F_{avg}$ \\
\citet{zhangmprf2025} & Test & GPT-3.5/4 & $F_{avg}$ \\
\citet{shi2026empirical} & Test & Multiple & Macro-F1 \\
\citet{nguyen2025external} & Test & GPT-4o-mini & $F_{avg}$ \\
\bottomrule
\end{tabular}
\caption{Inconsistencies across existing LLM-based stance detection studies.}
\label{tab:inconsistency}
\end{table}

We conduct a systematic comparison of LLM-based stance detection methods under strictly controlled conditions.
Our contributions are:

\begin{itemize}
    \item We build a unified evaluation covering four datasets, five methods, and 15 LLMs under a strictly controlled protocol, and release all per-target results.
    \item We provide the first systematic comparison between prompt-based and agent-based methods under identical conditions, showing that agent-based methods do not outperform the best prompt-based methods despite requiring significantly more computation.
    \item We analyze the effects of model scale and reasoning enhancement, finding that scaling gains plateau around 32B and that reasoning distillation does not consistently help stance detection.
\end{itemize}

%=============================================
\section{Related Work}
%=============================================

\subsection{Stance Detection with LLMs}

Prior to LLMs, stance detection relied on supervised deep learning models.
\citet{zhang2020sekt} proposed transferring semantic-emotion knowledge across targets.
\citet{zhang2023nps} introduced neural production systems for interpretable stance detection.
More recently, 
% \citet{zhang2024kai} proposed a knowledge-augmented interpretable network that uses LLM-elicited analysis perspectives for zero-shot stance detection, and 
\citet{dai2025llmltn} enhanced logic tensor networks with LLM-generated knowledge for improved stance reasoning.

With the emergence of instruction-following LLMs, researchers have explored prompt-based approaches.
\citet{zhang2022chatgpt} first explored using ChatGPT with a zero-shot prompt for stance detection.
\citet{zhang2022autocot} proposed Auto-CoT, which automatically generates chain-of-thought demonstrations through clustering.
\citet{zhang2023stsqa} introduced reasoning examples with human verification.

More complex approaches use multi-agent collaboration.
\citet{lan2024cola} proposed COLA, where expert agents analyze text and stance agents debate before a judge makes the final call.
\citet{zhangmprf2025} proposed MPRF, which uses four reasoning paths with quality scoring and refinement.
\citet{zhang2025logimdf} proposed LogiMDF, which fuses multiple LLMs' decision processes through first-order logic rules and hypergraph networks.
% \citet{zhang2025fuzzy} further explored fuzzy logic fusion of LLM predictions for stance detection.
In the zero-shot setting, \citet{ma2025cirf} proposed a cognitive inductive reasoning framework that abstracts transferable reasoning schemas from unlabeled text.

Each of these works reports results under different settings, making cross-paper comparison unreliable.

\subsection{Empirical Studies of LLM-based Stance Detection}

The most closely related work is \citet{shi2026empirical}, who conducted an empirical study of in-context learning methods for stance classification.
Their study compares several prompting strategies, including zero-shot, few-shot, and chain-of-thought variants, across multiple LLMs.
Our work differs in three key aspects.
First, we include agent-based methods (COLA, MPRF) that involve multi-agent debate, which their study does not cover; the comparison between prompt-based and agent-based paradigms is the central question of our evaluation.
Second, we cover a broader range of 15 LLMs from six model families with a systematic scale gradient from 7B to 72B+, while their study uses a smaller set of models.
Third, we report per-target results for all 14 subtasks across all model--method combinations, providing a more fine-grained analysis.

%=============================================
\section{Experimental Setup}
%=============================================

\subsection{Datasets}

We select four datasets that cover different domains, label sets, and difficulty levels (Table~\ref{tab:datasets}).

\begin{table*}[t]
\centering
\small
\begin{tabular}{llrll}
\toprule
\textbf{Dataset} & \textbf{Subtasks} & \textbf{Test Size} & \textbf{Labels} & \textbf{Domain} \\
\midrule
SemEval-2016 & DT, HC, FM, LA, A, CC & 208/294/278/280/220/169 & FAVOR, AGAINST, NONE & Social issues \\
P-Stance & Trump, Biden, Sanders & 777/745/635 & FAVOR, AGAINST & U.S. politics \\
COVID-19 & WA, SC, SH, AF & 200/200/200/200 & FAVOR, AGAINST, NONE & Public health \\
VAST & VAST & 1,460 & FAVOR, AGAINST, NONE & Diverse topics \\
\bottomrule
\end{tabular}
\caption{Datasets used in our evaluation. Total: 14 subtasks, 5,666 test samples. SemEval-2016 targets: DT=Donald Trump, HC=Hillary Clinton, FM=Feminist Movement, LA=Legalization of Abortion, A=Atheism, CC=Climate Change. COVID-19 targets: WA=Wearing Masks, SC=School Closures, SH=Stay at Home, AF=Anthony Fauci.}
\label{tab:datasets}
\end{table*}

\textbf{SemEval-2016 Task 6} \cite{mohammad2016semeval} is a classic tweet stance detection dataset with six social and political topics.

\textbf{P-Stance} \cite{li2021pstance} targets three U.S. political figures with binary labels (FAVOR, AGAINST).

\textbf{COVID-19-Stance} \cite{glandt2021covid} covers four pandemic-related targets with three labels.

\textbf{VAST} \cite{allaway2020vast} is a large-scale zero-shot dataset with diverse targets, considered the most challenging among the four.

These datasets are selected because they cover different domains, different label sets (binary and ternary), and different difficulty levels (in-target and zero-shot).
We use only the standard test split for all datasets.

\subsection{Methods}

We evaluate five methods in two categories (Table~\ref{tab:methods}).

\begin{table}[t]
\centering
\small
\begin{tabular}{lcrl}
\toprule
\textbf{Method} & \textbf{Cat.} & \textbf{Calls} & \textbf{Mechanism} \\
\midrule
Direct & P & 1 & Zero-shot query \\
Auto-CoT & P & 2+ & Auto chain-of-thought \\
StSQA & P & 1 & Reasoning example \\
\midrule
COLA & A & 7 & Role-infused debate \\
MPRF & A & 12+ & Multi-path reasoning \\
\bottomrule
\end{tabular}
\caption{Methods evaluated. Cat.: P = Prompt-based, A = Agent-based. Calls: LLM API calls per sample.}
\label{tab:methods}
\end{table}

\subsubsection{Prompt-based Methods}

\textbf{Direct Prompting} \cite{zhang2022chatgpt} uses a zero-shot prompt that asks the LLM to select a stance label for the given text and target. No examples or reasoning steps are provided.

\textbf{Auto-CoT} \cite{zhang2022autocot} first generates reasoning chains using zero-shot CoT, then clusters them, and selects one representative from each cluster as a demonstration for the final prompt.

\textbf{StSQA} \cite{zhang2023stsqa} selects one sample from the dataset, generates both stance and reasoning using an LLM, verifies the output manually, and uses it as a one-shot demonstration.

\subsubsection{Agent-based Methods}

\textbf{COLA} \cite{lan2024cola} uses three stages: (1) three expert agents (linguist, domain expert, social media analyst) analyze the text; (2) three stance agents argue for FAVOR, AGAINST, and NONE respectively; (3) a judge agent synthesizes all analyses and arguments. This requires about 7 LLM calls per sample.

\textbf{MPRF} \cite{zhangmprf2025} uses four reasoning paths (sentiment, factual, expert, public opinion), scores each on relevance, evidence strength, and logical consistency, refines low-scoring paths, and combines results through weighted voting. This requires 12+ LLM calls per sample.

\subsection{Models}

We select 15 LLMs from six families (Table~\ref{tab:models}).

\begin{table}[t]
\centering
\small
\begin{tabular}{lll}
\toprule
\textbf{Family} & \textbf{Model} & \textbf{Size} \\
\midrule
\multirow{2}{*}{OpenAI} & GPT-3.5-Turbo & closed \\
 & GPT-4o-mini & closed \\
\midrule
\multirow{3}{*}{DeepSeek} & DS-R1-Qwen-7B & 7B \\
 & DS-R1-Qwen-14B & 14B \\
 & DS-R1-Qwen-32B & 32B \\
\midrule
\multirow{3}{*}{Llama} & Llama-3.1-8B & 8B \\
 & Llama-3.1-70B & 70B \\
 & Llama-3.3-70B & 70B \\
\midrule
\multirow{4}{*}{Qwen} & Qwen2.5-7B & 7B \\
 & Qwen2.5-14B & 14B \\
 & Qwen2.5-32B & 32B \\
 & Qwen2.5-72B & 72B \\
\midrule
Claude & Claude-Haiku-4.5 & closed \\
\midrule
\multirow{2}{*}{Gemini} & Gemini-2.5-Flash & closed \\
 & Gemini-3-Flash & closed \\
\bottomrule
\end{tabular}
\caption{LLMs used. DS-R1-Qwen = DeepSeek-R1-Distill-Qwen. All open-source models use the instruct version.}
\label{tab:models}
\end{table}

Models are selected to provide: (1) a scale gradient from 7B to 72B+ through Qwen2.5 and DeepSeek families; (2) diversity across six families; (3) comparison between reasoning-enhanced models (DeepSeek-R1) and general models (Qwen2.5).

\subsection{Evaluation Protocol}

We use the standard test split for all datasets.
Temperature is set to 0 for deterministic outputs.
We extract stance labels from model outputs using keyword matching (FAVOR, AGAINST, NONE).
Following prior work \cite{mohammad2016semeval, lan2024cola}, we use macro-averaged F1 over the target classes: $F_{avg}$(FAVOR, AGAINST) for SemEval-2016 and P-Stance, and $F_{avg}$(FAVOR, AGAINST, NONE) for COVID-19 and VAST.
Some model--method combinations have incomplete results due to model refusal or API failure; we mark these cases explicitly and report the number of completed subtasks.

%=============================================
\section{Results}
%=============================================

\subsection{Overall Results}

Table~\ref{tab:overall} presents the overall results for all 15 models and five methods. The value in each cell is the average $F_{avg}$ across all 14 subtasks (or a subset if results are incomplete, as noted).

\begin{table*}[t]
\centering
\small
\begin{tabular}{l|ccccc}
\toprule
\textbf{Model} & \textbf{Direct} & \textbf{Auto-CoT} & \textbf{StSQA} & \textbf{COLA} & \textbf{MPRF} \\
\midrule
GPT-3.5-Turbo       & 53.6 & \textbf{76.3} & 66.3 & 62.2 & 61.0 \\
GPT-4o-mini          & 54.0 & \textbf{77.0} & 72.8 & 67.7 & 62.7 \\
DS-R1-Qwen-7B       & 49.9 & 58.5 & \textbf{60.1} & 55.8 & 19.7$^{\dagger}$ \\
DS-R1-Qwen-14B      & 65.2 & 69.3 & \textbf{73.2} & 67.8 & 64.1$^{\dagger}$ \\
DS-R1-Qwen-32B      & 62.8 & 75.1 & \textbf{75.4} & 68.5 & 59.9$^{\dagger}$ \\
Llama-3.1-8B        & 45.5 & \textbf{73.8} & 63.8 & 62.5 & 60.7 \\
Llama-3.1-70B       & 60.7 & \textbf{79.8} & 77.4 & 72.4 & 67.2 \\
Llama-3.3-70B       & 64.2 & \textbf{78.8} & 77.8 & 76.3 & 68.1 \\
Qwen2.5-7B          & 54.6 & \textbf{68.6} & 66.6 & 63.8 & 56.7$^{\dagger}$ \\
Qwen2.5-14B         & 63.0 & 74.2 & \textbf{76.3} & 71.4 & 67.7 \\
Qwen2.5-32B         & 66.3 & \textbf{76.4} & 76.0 & 73.3 & 71.8 \\
Qwen2.5-72B         & 67.9 & 72.5 & \textbf{76.7} & 72.3 & 65.8 \\
Claude-Haiku-4.5     & 66.3 & 64.6$^{\dagger}$ & \textbf{78.6} & 76.1 & 72.2 \\
Gemini-2.5-Flash     & 74.4 & 79.6 & \textbf{79.6} & 69.8 & 73.8 \\
Gemini-3-Flash       & 76.5 & 80.9 & \textbf{83.3} & 73.2 & 71.5 \\
\midrule
\textbf{Method Avg.} & 61.7 & 73.7 & \textbf{73.6} & 68.9 & 62.9 \\
\bottomrule
\end{tabular}
\caption{Overall results: average $F_{avg}$ (\%) across all 14 subtasks. Best result per model in bold. $^{\dagger}$Incomplete results (fewer than 14 subtasks completed); see Section~\ref{sec:incomplete} for details.}
\label{tab:overall}
\end{table*}

Several patterns emerge from these results.
First, Gemini-3-Flash achieves the highest scores across most methods. With StSQA, it reaches 83.3\%, the best overall result.
Second, Auto-CoT and StSQA are consistently the strongest methods, with overall averages of 73.7\% and 73.6\% respectively.
Third, Direct Prompting performs considerably worse than others, with an overall average of only 61.7\%.

\subsubsection{Incomplete Results}
\label{sec:incomplete}

Five model--method combinations have incomplete results.
MPRF on DeepSeek-R1-Distill-Qwen models has significant failures: 9/14 subtasks for the 7B and 14B models, and only 6/14 for the 32B model. The 7B model achieves only 19.7\% on completed subtasks, suggesting fundamental incompatibility between MPRF's multi-path reasoning pipeline and smaller reasoning-distilled models.
MPRF on Qwen2.5-7B completes 11/14 subtasks.
Auto-CoT on Claude-Haiku-4.5 completes 10/14 subtasks (missing four COVID-19 targets).
We report averages over completed subtasks only and mark all affected cells.

\subsection{Prompt-based vs. Agent-based}

\begin{figure}[t]
\centering
\includegraphics[width=\columnwidth]{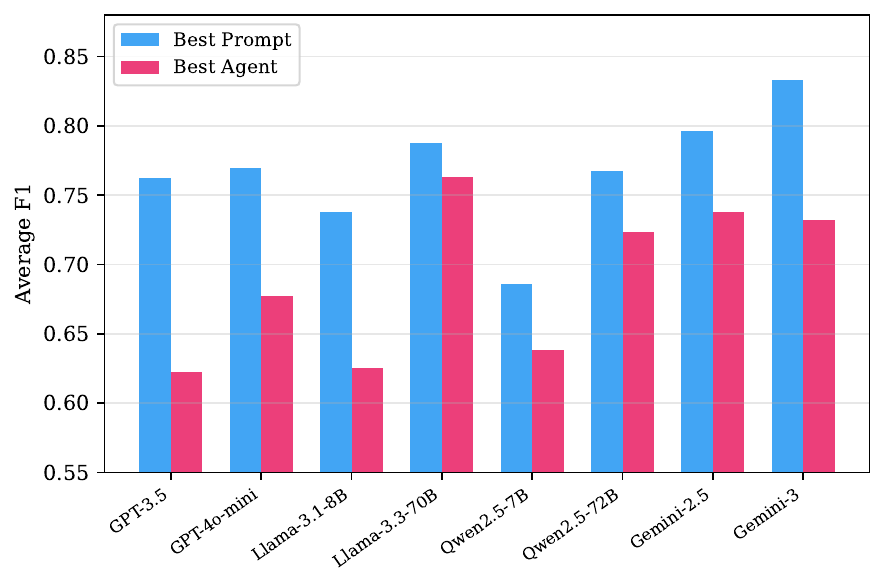}
\caption{Best prompt-based method vs. best agent-based method for each model. The best prompt method outperforms the best agent method on all models with complete results.}
\label{fig:prompt_vs_agent}
\end{figure}

A central question is whether multi-agent debate methods justify their additional computation cost.
Figure~\ref{fig:prompt_vs_agent} compares the best prompt-based method with the best agent-based method for each model.

On all models with complete results, the best prompt-based method outperforms the best agent-based method. The average gap is 6.5 percentage points.
On some models, the gap is especially large: 14.1 points on GPT-3.5-Turbo, 11.3 points on Llama-3.1-8B, and 10.1 points on Gemini-3-Flash.
The smallest gaps appear on Llama-3.3-70B (2.5 points) and Claude-Haiku-4.5 (2.5 points).

\begin{figure}[t]
\centering
\includegraphics[width=\columnwidth]{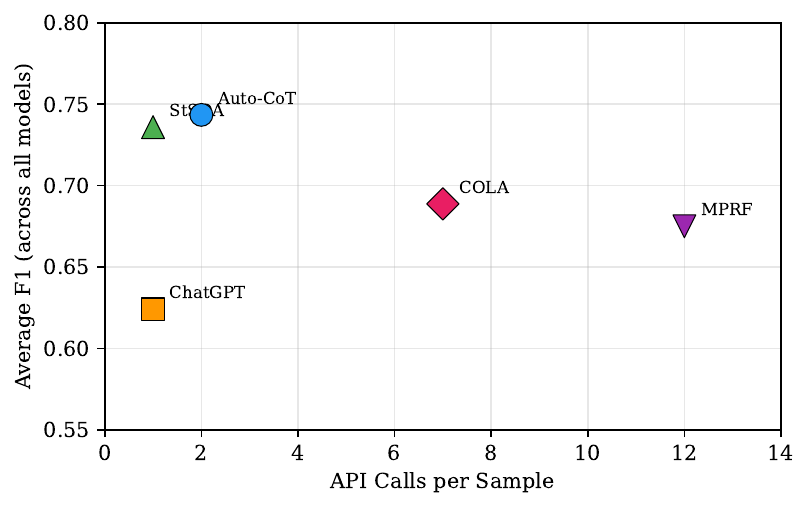}
\caption{Cost-performance trade-off. X-axis: API calls per sample. Y-axis: average F1 across all models. More expensive methods (COLA, MPRF) do not achieve higher performance.}
\label{fig:cost}
\end{figure}

Agent methods require 7 to 12 times more API calls per sample (Table~\ref{tab:methods}).
Figure~\ref{fig:cost} shows the cost-performance trade-off across all methods.

Auto-CoT achieves the highest average performance with only about 2 API calls per sample, while MPRF uses 12+ calls but scores lower.
StSQA is particularly efficient: it uses only 1 API call (with pre-computed demonstrations) and achieves comparable performance to Auto-CoT.

\subsection{Effect of Model Scale}

\begin{figure}[t]
\centering
\includegraphics[width=\columnwidth]{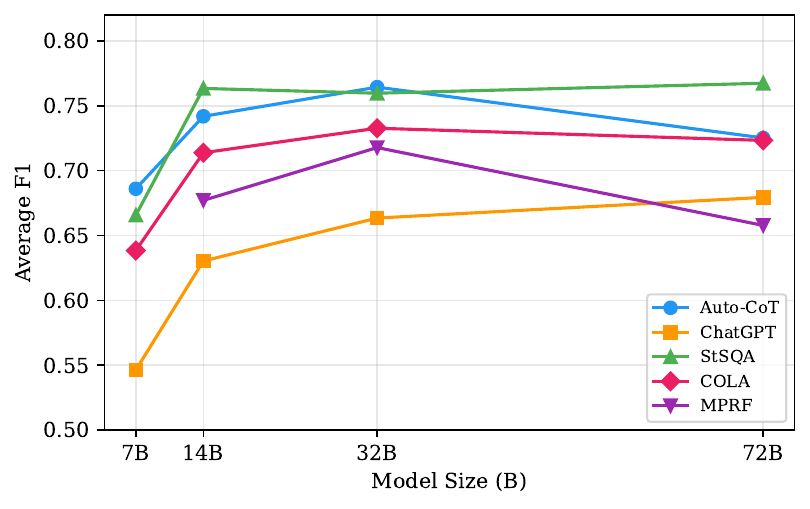}
\caption{Effect of model scale on performance using Qwen2.5 (7B, 14B, 32B, 72B). All methods benefit from scaling up to 32B, but gains diminish or reverse beyond that point.}
\label{fig:scaling}
\end{figure}

Figure~\ref{fig:scaling} shows how performance changes with model scale using the Qwen2.5 family (7B to 72B).

All methods benefit from scaling from 7B to 32B.
The improvement is substantial: Auto-CoT improves from 68.6\% to 76.4\% (+7.8 points), and StSQA improves from 66.6\% to 76.0\% (+9.4 points).
However, performance does not consistently improve from 32B to 72B.
For Auto-CoT, it actually decreases from 76.4\% to 72.5\%.
StSQA shows only a marginal change from 76.0\% to 76.7\%.

This suggests that for stance detection, the gains from scaling plateau around 32B.
The improvement from upgrading the method (e.g., Direct to Auto-CoT) is comparable to the improvement from scaling the model (e.g., 7B to 32B), both around 8 to 10 percentage points.

\subsection{Reasoning-Enhanced Models}

\begin{figure}[t]
\centering
\includegraphics[width=\columnwidth]{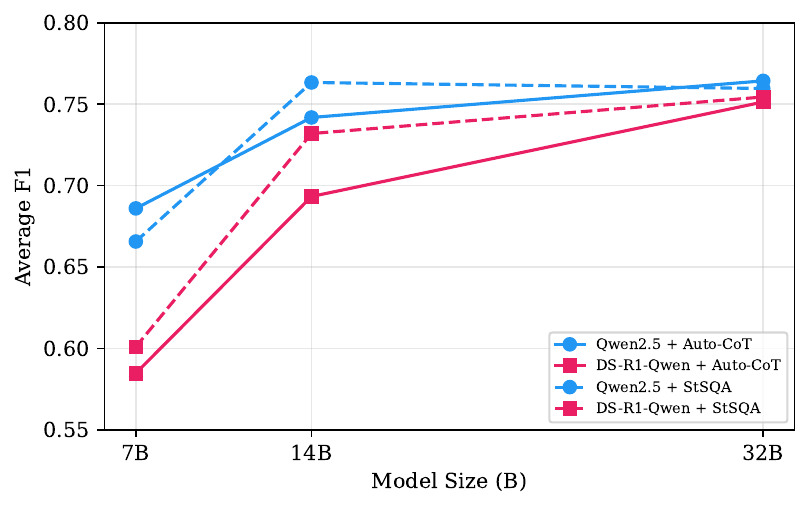}
\caption{Reasoning-enhanced models (DS-R1-Qwen) vs. general models (Qwen2.5) at matched sizes.}
\label{fig:reasoning}
\end{figure}

Figure~\ref{fig:reasoning} compares DeepSeek-R1-Distill-Qwen (reasoning-enhanced) with Qwen2.5 (general) at matched parameter sizes.

At 7B, DeepSeek-R1 is weaker than Qwen2.5 on both Auto-CoT (58.5\% vs. 68.6\%) and StSQA (60.1\% vs. 66.6\%).
At 14B, the gap narrows but Qwen2.5 still leads on Auto-CoT (74.2\% vs. 69.3\%).
At 32B, the two converge: DeepSeek-R1 reaches 75.1\% on Auto-CoT versus 76.4\% for Qwen2.5.

These results suggest that reasoning distillation does not transfer well to stance detection at smaller scales.
Stance detection may rely more on world knowledge and social understanding than on step-by-step reasoning, which explains why general instruction-following models perform comparably or better.

\subsection{Per-Target Results}

We present the complete per-target results for each method in Tables~\ref{tab:detail_direct}--\ref{tab:detail_mprf}. These tables reveal substantial variation across targets that is hidden in aggregate scores.

\subsubsection{SemEval-2016}

Among the six targets, Atheism (A) is consistently the hardest for Direct Prompting, with most models scoring below 55\% (GPT-3.5: 33.2\%, GPT-4o-mini: 31.6\%, Llama-3.1-8B: 41.3\%). This target substantially improves with Auto-CoT and StSQA, suggesting that reasoning demonstrations are especially helpful for implicit stance expressions.
Hillary Clinton (HC) is the easiest, with most methods achieving above 75\%.

\subsubsection{P-Stance}

P-Stance is the easiest dataset overall, with most methods scoring above 75\%. This is expected because P-Stance uses binary labels and covers well-defined political targets.
COLA achieves its best relative performance on this dataset. On Llama-3.3-70B, COLA reaches an average of 86.5\% across P-Stance targets, the highest agent-method score on any dataset--model combination. This suggests that political stance detection may benefit from the multi-perspective analysis that COLA provides.

\subsubsection{VAST}

VAST is the zero-shot dataset with diverse unseen targets. Direct Prompting struggles severely (GPT-3.5: 43.7\%, Llama-3.1-8B: 35.1\%), but Auto-CoT and StSQA bring substantial improvements (Gemini-3-Flash StSQA: 88.0\%).

% ===========================================================
% DETAILED PER-TARGET TABLES
% ===========================================================

\begin{table*}[!ht]
\centering
\scriptsize
\setlength{\tabcolsep}{3pt}
\begin{tabular}{l|cccccc|ccc|c|cccc|c}
\toprule
& \multicolumn{6}{c|}{\textbf{SemEval-2016}} & \multicolumn{3}{c|}{\textbf{P-Stance}} & \textbf{VAST} & \multicolumn{4}{c|}{\textbf{COVID-19}} & \\
\textbf{Model} & DT & HC & FM & LA & A & CC & Trp & Bid & San & VAST & WA & SC & AF & SH & \textbf{Avg} \\
\midrule
GPT-3.5        & 65.9 & 67.4 & 62.7 & 55.2 & 33.2 & 49.1 & 65.2 & 76.7 & 70.4 & 43.7 & 44.0 & 25.5 & 37.5 & 53.5 & 53.6 \\
GPT-4o-mini    & 67.1 & 74.5 & 64.8 & 53.3 & 31.6 & 49.7 & 66.1 & 77.1 & 71.0 & 43.7 & 47.9 & 21.9 & 33.5 & 54.0 & 54.0 \\
DS-R1-7B       & 55.4 & 56.6 & 58.1 & 46.5 & 28.7 & 46.2 & 62.7 & 75.5 & 65.6 & 50.9 & 47.6 & 26.7 & 39.6 & 38.6 & 49.9 \\
DS-R1-14B      & 67.4 & 77.3 & 73.7 & 67.7 & 59.9 & 56.0 & 74.4 & 81.0 & 76.3 & 64.3 & 61.3 & 39.0 & 52.6 & 62.5 & 65.2 \\
DS-R1-32B      & 54.3 & 66.4 & 63.4 & 54.7 & 58.8 & 67.8 & 79.0 & 81.9 & 78.2 & 63.3 & 61.5 & 43.4 & 53.5 & 52.7 & 62.8 \\
Llama-8B       & 47.7 & 68.3 & 56.9 & 56.5 & 41.3 & 45.8 & 50.6 & 70.0 & 64.7 & 35.1 & 31.6 & 13.7 & 27.1 & 27.7 & 45.5 \\
Llama-70B      & 71.4 & 78.8 & 72.5 & 70.5 & 52.2 & 59.2 & 77.0 & 82.6 & 77.2 & 49.3 & 52.2 & 25.1 & 47.0 & 34.6 & 60.7 \\
Llama3.3-70B   & 72.0 & 81.6 & 75.7 & 68.8 & 53.7 & 61.3 & 78.1 & 84.1 & 78.7 & 50.6 & 58.2 & 29.7 & 49.4 & 56.5 & 64.2 \\
Qwen-7B        & 65.1 & 66.8 & 64.6 & 54.0 & 37.9 & 61.0 & 59.1 & 76.3 & 69.2 & 48.9 & 54.7 & 31.4 & 44.6 & 31.4 & 54.6 \\
Qwen-14B       & 68.6 & 77.3 & 73.7 & 64.5 & 56.4 & 66.1 & 69.5 & 81.0 & 75.5 & 52.7 & 56.0 & 35.5 & 43.1 & 62.6 & 63.0 \\
Qwen-32B       & 69.7 & 73.4 & 75.2 & 63.5 & 55.8 & 72.1 & 74.7 & 83.1 & 77.6 & 58.2 & 64.5 & 41.6 & 57.3 & 62.0 & 66.3 \\
Qwen-72B       & 71.9 & 78.4 & 73.7 & 69.1 & 55.8 & 73.8 & 77.2 & 83.2 & 76.8 & 62.1 & 69.0 & 34.9 & 57.5 & 67.8 & 67.9 \\
Claude-H4.5    & 70.8 & 77.0 & 76.7 & 70.4 & 38.8 & 73.2 & 82.5 & 82.3 & 77.3 & 55.3 & 65.1 & 36.1 & 59.4 & 63.7 & 66.3 \\
Gem-2.5F       & 67.6 & 80.5 & 74.7 & 67.6 & 74.6 & 81.2 & 87.8 & 84.7 & 81.3 & 73.2 & 74.9 & 59.2 & 70.9 & 63.1 & 74.4 \\
Gem-3F         & 66.2 & 78.2 & 76.6 & 67.8 & 75.8 & 79.1 & 91.8 & 86.0 & 83.9 & 78.0 & 77.0 & 69.7 & 67.1 & 74.3 & 76.5 \\
\bottomrule
\end{tabular}
\caption{Per-target results (\%) for \textbf{Direct Prompting}.}
\label{tab:detail_direct}
\end{table*}

\begin{table*}[!ht]
\centering
\scriptsize
\setlength{\tabcolsep}{3pt}
\begin{tabular}{l|cccccc|ccc|c|cccc|c}
\toprule
& \multicolumn{6}{c|}{\textbf{SemEval-2016}} & \multicolumn{3}{c|}{\textbf{P-Stance}} & \textbf{VAST} & \multicolumn{4}{c|}{\textbf{COVID-19}} & \\
\textbf{Model} & DT & HC & FM & LA & A & CC & Trp & Bid & San & VAST & WA & SC & AF & SH & \textbf{Avg} \\
\midrule
GPT-3.5        & 73.2 & 89.3 & 77.5 & 68.5 & 76.5 & 74.2 & 84.0 & 79.9 & 80.0 & 78.3 & 77.1 & 63.1 & 66.3 & 79.9 & 76.3 \\
GPT-4o-mini    & 71.2 & 87.1 & 74.6 & 69.6 & 74.4 & 80.0 & 85.2 & 79.3 & 79.0 & 81.1 & 77.1 & 66.7 & 70.6 & 81.9 & 77.0 \\
DS-R1-7B       & 65.3 & 68.7 & 56.6 & 57.4 & 39.2 & 57.3 & 63.0 & 68.7 & 63.5 & 59.6 & 54.7 & 40.1 & 61.3 & 63.3 & 58.5 \\
DS-R1-14B      & 61.8 & 83.8 & 68.7 & 45.0 & 72.8 & 63.2 & 84.1 & 81.2 & 77.9 & 78.4 & 69.2 & 55.7 & 73.3 & 55.7 & 69.3 \\
DS-R1-32B      & 66.5 & 84.2 & 69.0 & 66.9 & 64.9 & 82.3 & 86.3 & 82.7 & 81.2 & 78.4 & 73.5 & 70.6 & 74.8 & 70.6 & 75.1 \\
Llama-8B       & 68.8 & 77.7 & 72.1 & 71.3 & 68.2 & 82.4 & 83.6 & 80.7 & 79.0 & 76.5 & 74.4 & 59.0 & 61.9 & 77.6 & 73.8 \\
Llama-70B      & 67.5 & 85.0 & 74.6 & 65.2 & 74.4 & 88.4 & 90.1 & 83.6 & 80.8 & 82.2 & 83.4 & 75.0 & 85.2 & 82.4 & 79.8 \\
Llama3.3-70B   & 71.1 & 86.1 & 76.2 & 65.8 & 74.7 & 84.8 & 87.5 & 83.8 & 81.8 & 78.6 & 82.5 & 68.9 & 76.6 & 84.2 & 78.8 \\
Qwen-7B        & 69.3 & 79.2 & 69.7 & 65.2 & 59.0 & 57.7 & 83.2 & 79.2 & 77.0 & 78.8 & 65.2 & 46.2 & 74.7 & 56.1 & 68.6 \\
Qwen-14B       & 63.9 & 80.8 & 76.8 & 68.0 & 74.8 & 80.1 & 83.6 & 82.2 & 79.2 & 79.6 & 75.4 & 64.5 & 72.3 & 57.2 & 74.2 \\
Qwen-32B       & 68.7 & 75.9 & 76.0 & 66.9 & 70.4 & 83.6 & 88.3 & 81.6 & 81.9 & 61.7 & 81.8 & 77.9 & 73.7 & 81.8 & 76.4 \\
Qwen-72B       & 68.2 & 85.3 & 75.3 & 70.6 & 37.1 & 67.8 & 90.5 & 84.0 & 82.0 & 61.7 & 74.4 & 69.9 & 76.3 & 72.2 & 72.5 \\
Claude-H4.5    & 70.8 & 80.7 & 73.5 & 40.2 & 68.7 & 66.8 & 43.9 & 43.8 & 77.4 & 80.7 &  --  &  --  &  --  &  --  & 64.6$^{\dagger}$ \\
Gem-2.5F       & 68.8 & 85.6 & 72.4 & 65.9 & 75.5 & 81.5 & 91.2 & 84.3 & 81.8 & 81.3 & 86.5 & 77.2 & 80.5 & 82.1 & 79.6 \\
Gem-3F         & 73.1 & 84.8 & 75.6 & 66.7 & 77.1 & 79.5 & 90.4 & 84.6 & 82.4 & 84.1 & 82.8 & 84.9 & 84.2 & 82.2 & 80.9 \\
\bottomrule
\end{tabular}
\caption{Per-target results (\%) for \textbf{Auto-CoT}. $^{\dagger}$10/14 subtasks completed.}
\label{tab:detail_autocot}
\end{table*}

\begin{table*}[!ht]
\centering
\scriptsize
\setlength{\tabcolsep}{3pt}
\begin{tabular}{l|cccccc|ccc|c|cccc|c}
\toprule
& \multicolumn{6}{c|}{\textbf{SemEval-2016}} & \multicolumn{3}{c|}{\textbf{P-Stance}} & \textbf{VAST} & \multicolumn{4}{c|}{\textbf{COVID-19}} & \\
\textbf{Model} & DT & HC & FM & LA & A & CC & Trp & Bid & San & VAST & WA & SC & AF & SH & \textbf{Avg} \\
\midrule
GPT-3.5        & 70.4 & 79.0 & 73.9 & 69.6 & 68.6 & 56.5 & 79.5 & 83.5 & 75.3 & 54.1 & 60.2 & 33.6 & 54.8 & 69.4 & 66.3 \\
GPT-4o-mini    & 73.4 & 84.4 & 71.2 & 68.2 & 48.1 & 73.9 & 78.2 & 83.7 & 76.1 & 71.4 & 71.4 & 69.3 & 72.1 & 77.2 & 72.8 \\
DS-R1-7B       & 61.8 & 72.0 & 56.9 & 45.4 & 40.0 & 59.2 & 62.5 & 77.3 & 70.9 & 65.0 & 61.4 & 40.4 & 67.5 & 61.0 & 60.1 \\
DS-R1-14B      & 67.3 & 83.2 & 66.5 & 67.2 & 61.7 & 75.4 & 78.8 & 86.0 & 78.8 & 75.8 & 75.0 & 55.8 & 75.3 & 78.0 & 73.2 \\
DS-R1-32B      & 72.1 & 84.1 & 72.2 & 65.0 & 61.2 & 79.7 & 83.4 & 84.8 & 81.0 & 78.2 & 77.1 & 63.8 & 75.3 & 78.3 & 75.4 \\
Llama-8B       & 68.7 & 78.1 & 70.8 & 64.6 & 51.4 & 63.2 & 73.7 & 83.5 & 74.2 & 57.2 & 55.4 & 35.8 & 50.3 & 66.2 & 63.8 \\
Llama-70B      & 72.9 & 87.5 & 75.2 & 69.0 & 75.4 & 85.6 & 88.5 & 86.6 & 82.2 & 69.7 & 65.2 & 66.5 & 80.0 & 79.9 & 77.4 \\
Llama3.3-70B   & 72.2 & 86.8 & 74.5 & 67.2 & 73.9 & 89.7 & 86.9 & 85.6 & 82.3 & 73.5 & 73.5 & 66.1 & 77.0 & 80.3 & 77.8 \\
Qwen-7B        & 67.0 & 80.3 & 68.5 & 61.4 & 48.6 & 77.5 & 72.6 & 83.3 & 72.6 & 64.8 & 60.5 & 41.5 & 63.0 & 70.2 & 66.6 \\
Qwen-14B       & 71.6 & 81.8 & 73.0 & 69.3 & 73.3 & 86.9 & 81.7 & 86.9 & 79.9 & 75.3 & 76.8 & 59.0 & 74.7 & 78.3 & 76.3 \\
Qwen-32B       & 75.1 & 85.7 & 70.3 & 67.5 & 67.2 & 80.0 & 87.4 & 85.7 & 81.7 & 78.1 & 74.0 & 52.5 & 77.8 & 80.4 & 76.0 \\
Qwen-72B       & 70.1 & 85.6 & 79.4 & 68.5 & 70.9 & 64.6 & 87.3 & 85.4 & 82.6 & 79.0 & 79.6 & 63.6 & 79.3 & 78.5 & 76.7 \\
Claude-H4.5    & 70.9 & 85.2 & 75.6 & 68.6 & 80.0 & 77.3 & 88.0 & 86.1 & 83.2 & 78.7 & 76.0 & 74.0 & 74.8 & 81.4 & 78.6 \\
Gem-2.5F       & 70.8 & 88.3 & 78.5 & 68.7 & 62.1 & 82.5 & 92.1 & 87.7 & 83.9 & 82.7 & 80.7 & 79.7 & 80.3 & 76.7 & 79.6 \\
Gem-3F         & 76.3 & 85.4 & 79.5 & 70.8 & 80.2 & 80.0 & 93.3 & 88.2 & 84.3 & 88.0 & 88.2 & 84.6 & 81.9 & 85.5 & 83.3 \\
\bottomrule
\end{tabular}
\caption{Per-target results (\%) for \textbf{StSQA}.}
\label{tab:detail_stsqa}
\end{table*}

\begin{table*}[!ht]
\centering
\scriptsize
\setlength{\tabcolsep}{3pt}
\begin{tabular}{l|cccccc|ccc|c|cccc|c}
\toprule
& \multicolumn{6}{c|}{\textbf{SemEval-2016}} & \multicolumn{3}{c|}{\textbf{P-Stance}} & \textbf{VAST} & \multicolumn{4}{c|}{\textbf{COVID-19}} & \\
\textbf{Model} & DT & HC & FM & LA & A & CC & Trp & Bid & San & VAST & WA & SC & AF & SH & \textbf{Avg} \\
\midrule
GPT-3.5        & 66.2 & 72.0 & 60.1 & 50.6 & 59.6 & 79.5 & 85.9 & 83.5 & 82.0 & 50.9 & 55.1 & 33.1 & 50.8 & 42.2 & 62.2 \\
GPT-4o-mini    & 65.0 & 78.1 & 65.6 & 63.3 & 61.1 & 78.2 & 88.2 & 84.3 & 82.2 & 65.3 & 60.8 & 43.5 & 56.7 & 55.7 & 67.7 \\
DS-R1-7B       & 60.8 & 63.2 & 57.9 & 47.1 & 41.0 & 49.2 & 64.6 & 69.9 & 67.5 & 60.1 & 59.1 & 33.7 & 54.6 & 53.0 & 55.8 \\
DS-R1-14B      & 66.9 & 76.6 & 61.9 & 64.9 & 60.8 & 72.0 & 83.8 & 85.3 & 80.4 & 70.7 & 68.0 & 43.1 & 60.0 & 54.6 & 67.8 \\
DS-R1-32B      & 64.0 & 73.8 & 62.6 & 62.5 & 66.0 & 77.4 & 85.2 & 84.2 & 79.7 & 71.0 & 67.8 & 50.7 & 60.6 & 53.8 & 68.5 \\
Llama-8B       & 65.6 & 80.9 & 67.9 & 58.5 & 32.9 & 46.4 & 74.8 & 78.0 & 74.9 & 65.8 & 60.5 & 47.3 & 58.1 & 63.6 & 62.5 \\
Llama-70B      & 65.9 & 76.5 & 66.2 & 64.2 & 72.6 & 82.3 & 89.3 & 86.2 & 82.6 & 77.6 & 75.3 & 55.2 & 47.0 & 72.5 & 72.4 \\
Llama3.3-70B   & 68.6 & 81.2 & 67.9 & 67.3 & 73.4 & 87.3 & 89.4 & 86.8 & 83.2 & 77.1 & 78.8 & 57.6 & 70.0 & 79.5 & 76.3 \\
Qwen-7B        & 68.1 & 72.1 & 63.0 & 57.8 & 59.4 & 77.4 & 74.4 & 82.0 & 76.0 & 64.0 & 65.9 & 32.2 & 52.9 & 48.5 & 63.8 \\
Qwen-14B       & 67.6 & 78.0 & 66.1 & 64.3 & 66.0 & 79.1 & 83.9 & 86.2 & 78.6 & 73.1 & 77.4 & 57.7 & 61.8 & 59.6 & 71.4 \\
Qwen-32B       & 66.7 & 78.1 & 68.3 & 66.0 & 68.4 & 76.4 & 87.4 & 86.9 & 82.3 & 77.6 & 72.9 & 59.6 & 65.4 & 69.7 & 73.3 \\
Qwen-72B       & 64.1 & 80.9 & 69.6 & 67.9 & 63.2 & 87.0 & 89.2 & 87.7 & 82.3 & 72.9 & 76.1 & 48.2 & 63.0 & 60.5 & 72.3 \\
Claude-H4.5    & 64.8 & 81.4 & 73.3 & 68.7 & 70.0 & 79.2 & 89.9 & 86.0 & 84.0 & 78.6 & 79.4 & 58.4 & 71.0 & 80.5 & 76.1 \\
Gem-2.5F       & 61.9 & 75.8 & 68.7 & 64.8 & 71.8 & 74.6 & 85.4 & 91.3 & 82.8 & 74.8 & 63.8 & 54.4 & 57.5 & 49.3 & 69.8 \\
Gem-3F         & 60.2 & 72.5 & 72.8 & 65.6 & 74.7 & 74.4 & 91.8 & 85.8 & 83.0 & 77.0 & 70.7 & 67.4 & 55.1 & 73.9 & 73.2 \\
\bottomrule
\end{tabular}
\caption{Per-target results (\%) for \textbf{COLA}.}
\label{tab:detail_cola}
\end{table*}

\begin{table*}[!ht]
\centering
\scriptsize
\setlength{\tabcolsep}{3pt}
\begin{tabular}{l|cccccc|ccc|c|cccc|c}
\toprule
& \multicolumn{6}{c|}{\textbf{SemEval-2016}} & \multicolumn{3}{c|}{\textbf{P-Stance}} & \textbf{VAST} & \multicolumn{4}{c|}{\textbf{COVID-19}} & \\
\textbf{Model} & DT & HC & FM & LA & A & CC & Trp & Bid & San & VAST & WA & SC & AF & SH & \textbf{Avg} \\
\midrule
GPT-3.5        & 64.6 & 71.4 & 71.1 & 64.5 & 44.5 & 57.3 & 66.8 & 83.1 & 78.4 & 43.5 & 58.0 & 34.0 & 53.0 & 64.3 & 61.0 \\
GPT-4o-mini    & 60.8 & 73.3 & 67.3 & 63.8 & 46.1 & 56.6 & 80.9 & 83.0 & 79.4 & 59.1 & 58.9 & 31.6 & 52.2 & 64.4 & 62.7 \\
DS-R1-7B       & 20.0 & 29.5 & 20.7 & 19.4 & 13.0 & 13.6 & 18.1 & 23.8 & 19.4 & --   & --   & --   & --   & --   & 19.7$^{\dagger}$ \\
DS-R1-14B      & 62.3 & 72.7 & 61.3 & 67.0 & 54.8 & 34.8 & 81.1 & 63.3 & 79.5 & --   & --   & --   & --   & --   & 64.1$^{\dagger}$ \\
DS-R1-32B      & 58.5 & 74.8 & 65.7 & 65.5 & 56.9 & 38.2 & --   & --   & --   & --   & --   & --   & --   & --   & 59.9$^{\dagger}$ \\
Llama-8B       & 67.6 & 74.3 & 66.8 & 63.5 & 48.6 & 51.7 & 75.7 & 81.7 & 75.4 & 48.9 & 51.5 & 31.7 & 51.0 & 61.9 & 60.7 \\
Llama-70B      & 52.8 & 68.4 & 64.7 & 69.0 & 58.8 & 68.1 & 84.7 & 84.5 & 80.1 & 69.6 & 75.7 & 25.7 & 71.3 & 67.6 & 67.2 \\
Llama3.3-70B   & 70.7 & 84.9 & 75.6 & 72.0 & 55.3 & 62.8 & 81.8 & 83.7 & 80.3 & 69.4 & 57.9 & 26.7 & 69.1 & 63.0 & 68.1 \\
Qwen-7B        & 65.5 & 79.2 & 68.0 & 60.9 & 31.8 & 58.1 & --   & --   & --   & 61.9 & 62.8 & 25.3 & 60.4 & 49.4 & 56.7$^{\dagger}$ \\
Qwen-14B       & 70.0 & 81.7 & 69.3 & 61.5 & 63.4 & 73.1 & 78.3 & 84.2 & 79.0 & 66.1 & 71.0 & 23.5 & 69.3 & 57.5 & 67.7 \\
Qwen-32B       & 70.4 & 85.8 & 77.1 & 70.5 & 71.4 & 70.3 & 84.1 & 84.6 & 81.1 & 65.9 & 75.9 & 20.5 & 72.6 & 74.8 & 71.8 \\
Qwen-72B       & 71.1 & 83.9 & 76.4 & 62.6 & 44.9 & 60.9 & 80.7 & 83.0 & 79.7 & 63.6 & 64.3 & 20.9 & 66.1 & 62.8 & 65.8 \\
Claude-H4.5    & 62.3 & 79.0 & 74.3 & 70.2 & 73.2 & 76.6 & 85.4 & 78.2 & 83.7 & 69.4 & 83.9 & 34.1 & 64.6 & 75.9 & 72.2 \\
Gem-2.5F       & 63.1 & 75.4 & 71.2 & 65.1 & 70.6 & 70.8 & 91.0 & 85.6 & 82.9 & 80.6 & 72.8 & 64.1 & 65.9 & 74.1 & 73.8 \\
Gem-3F         & 61.4 & 73.6 & 73.3 & 67.0 & 73.6 & 74.8 & 50.9 & 83.5 & 62.1 & 79.9 & 73.5 & 78.3 & 67.7 & 81.0 & 71.5 \\
\bottomrule
\end{tabular}
\caption{Per-target results (\%) for \textbf{MPRF}. $^{\dagger}$Incomplete: DS-R1-7B and DS-R1-14B complete 9/14, DS-R1-32B completes 6/14, Qwen-7B completes 11/14 subtasks. ``--'' = model failure.}
\label{tab:detail_mprf}
\end{table*}

\subsubsection{COVID-19}

COVID-19 shows the largest variance across targets. School Closures (SC) is particularly challenging: Direct Prompting scores range from 13.7\% (Llama-3.1-8B) to 69.7\% (Gemini-3-Flash). Even the best agent method COLA achieves only 33.1\% on SC with GPT-3.5. In contrast, Wearing Masks (WA) and Stay at Home (SH) are much easier, with several models exceeding 80\% using Auto-CoT.

\subsection{Cost-Performance Trade-off}

Table~\ref{tab:cost} summarizes the efficiency of each method.

\begin{table}[t]
\centering
\small
\begin{tabular}{lrr}
\toprule
\textbf{Method} & \textbf{Calls/Sample} & \textbf{Avg. F1 (\%)} \\
\midrule
Direct & 1 & 61.7 \\
StSQA & 1 & 73.6 \\
Auto-CoT & 2+ & 73.7 \\
COLA & 7 & 68.9 \\
MPRF & 12+ & 62.9 \\
\bottomrule
\end{tabular}
\caption{Cost-performance trade-off. StSQA achieves high performance with minimal API usage.}
\label{tab:cost}
\end{table}

StSQA offers the best efficiency: 73.6\% with a single inference call (demonstrations are pre-computed).
Auto-CoT achieves the highest absolute score (73.7\%) but requires 2+ calls per sample for clustering and demonstration construction.
COLA and MPRF have the worst efficiency, achieving lower scores with far more computation.
Notably, MPRF averages only 62.9\% across all models---lower even than Direct Prompting on some models---while consuming 12$\times$ more API calls.

%=============================================
\section{Discussion}
%=============================================

\subsection{Key Findings}

\textbf{Finding 1: Agent-based methods do not outperform the best prompt-based methods.}
On all models with complete results, the best prompt method (Auto-CoT or StSQA) outperforms the best agent method (COLA or MPRF) by an average of 6.5 points, despite agent methods using 7 to 12 times more API calls.

\textbf{Finding 2: Auto-CoT and StSQA are the most effective methods overall.}
Auto-CoT achieves the highest average (73.7\%) across all models, closely followed by StSQA (73.6\%). Auto-CoT benefits from diverse demonstrations selected through clustering, while StSQA benefits from explicit reasoning examples.

\textbf{Finding 3: Model scale matters more than method complexity up to 32B.}
Scaling from 7B to 32B improves performance by 8 to 10 points for the same method, comparable to the improvement from switching methods (e.g., Direct to Auto-CoT). Beyond 32B, gains diminish.

\textbf{Finding 4: Zero-shot Direct Prompting is unreliable.}
Direct Prompting scores only 61.7\% on average, 12 points below Auto-CoT and StSQA. Adding reasoning demonstrations consistently helps, with especially large gains on difficult targets like Atheism and VAST.

\textbf{Finding 5: Reasoning-enhanced models do not help at small scales.}
DeepSeek-R1-Distill-Qwen-7B underperforms Qwen2.5-7B by over 8 points on average. The gap narrows at larger scales but Qwen2.5 remains competitive.

\textbf{Finding 6: MPRF is fragile across model families.}
MPRF shows high failure rates on DeepSeek-R1 models (completing only 6--9 out of 14 subtasks) and produces anomalously low scores (19.7\% on DS-R1-7B). Its multi-path pipeline appears to amplify errors in smaller or reasoning-distilled models.

\subsection{Practical Recommendations}

Based on our results, we offer the following suggestions.
If cost is the primary concern, StSQA with a 32B+ model provides strong performance with minimal API usage.
If accuracy is the priority, Auto-CoT with Gemini-3-Flash or a 70B-class model achieves the best results.
Agent-based methods are not recommended for general stance detection, though COLA may help on political topics where multi-perspective analysis is valuable.

%=============================================
\section{Conclusion}
%=============================================

We present a systematic comparison of five LLM-based stance detection methods across 4 datasets, 14 subtasks, and 15 LLMs under a strictly controlled protocol.
The key finding is that multi-agent debate methods do not outperform well-designed prompting strategies, despite using significantly more computation.
Auto-CoT and StSQA emerge as the most effective approaches, and model scale provides consistent improvements up to 32B.

We release all code, prompts, and detailed per-target results to support future research and to serve as a reference for evaluating new stance detection methods.

\textbf{Limitations.}
All datasets are in English, and results may differ for other languages; recent efforts such as \citet{niu2025cmtcsd} have begun constructing Chinese stance detection resources, and extending our evaluation to multilingual settings is an important direction.
We do not include supervised fine-tuning baselines.
Closed-source models may update over time, affecting reproducibility.
Some model--method combinations have incomplete results due to model refusal or API limitations; these are explicitly marked throughout.

\bibliography{references}

\appendix

\section{Prompt Templates}
\label{app:prompts}

\subsection{Direct Prompting}

\begin{quote}
\small
\texttt{What's the attitude of the sentence: \{text\} to the target \{target\}. select from ``'AGAINST', 'FAVOR' or 'NONE'''.\\ Do NOT explain.\\ Do NOT add any other words.\\ Do NOT add punctuation or symbols.\\ Only select one stance.}
\end{quote}

\subsection{Auto-CoT}

\textbf{Step 1: Zero-shot CoT.}
\begin{quote}
\small
\texttt{Q: Based on the tweet: ``\{tweet\}'', determine the user's attitude towards ``\{target\}''. Options: against, none, favor.\\
A: Let's think step by step.}
\end{quote}

\textbf{Step 2:} Cluster reasoning chains and select one representative per cluster.

\textbf{Step 3: Inference with demonstrations.}
\begin{quote}
\small
\texttt{\{demo\_text\}\\
\\
Q: Based on the tweet: ``\{tweet\}'', determine the user's attitude towards ``\{target\}''. Options: against, none, favor.\\
A: Let's think step by step.}
\end{quote}

\subsection{StSQA}

\textbf{Inference prompt (with pre-computed example):}
\begin{quote}
\small
\texttt{You are an expert in stance detection.\\
Analyze the attitude of the given sentence to the target. Only select from 'AGAINST', 'FAVOR', 'NONE'.\\
Output your response STRICTLY as a valid JSON object.\\
Example:\\
Target: ``\{target\}''\\
Sentence: ``\{text\}''\\
Output:\\
\{``Reason'': ``\{reason\}'', ``Final Stance'': ``\{stance\}''\}\\
Now, analyze the following:\\
Target: ``\{target\}''\\
Sentence: ``\{text\}''}
\end{quote}

\subsection{COLA}

\textbf{Stage 1:} Three expert prompts (linguist, domain expert, social media analyst) each analyze the text from their perspective.

\textbf{Stage 2:} Three stance agents each argue for FAVOR, AGAINST, and NONE using the expert analyses.

\textbf{Stage 3: Judge prompt:}
\begin{quote}
\small
\texttt{Determine whether the sentence is in favor of or against \{target\}, or is irrelevant.\\
Sentence: \{tweet\}\\
Arguments in favor: \{favor\_response\}\\
Arguments against: \{against\_response\}\\
Choose from: AGAINST, FAVOR, NONE\\
Answer with only the option that is most accurate.}
\end{quote}

\subsection{MPRF}

\textbf{Four reasoning paths} (sentiment, factual, expert, public) each generate a stance prediction with explanation. Each path is scored (1-10) on relevance, evidence strength, and logical consistency. Low-scoring paths are refined. The final decision uses weighted fusion and majority voting. Full templates are in our code repository.

\end{document}